%% file: main.tex
\documentclass[10pt,twocolumn,letterpaper]{article}

\usepackage{cvpr}              
\usepackage{float}
\usepackage{graphicx}
\usepackage{caption}
\usepackage{subcaption}
\usepackage{array}
\input{preamble}
\definecolor{cvprblue}{rgb}{0.21,0.49,0.74}
\usepackage[pagebackref,breaklinks,colorlinks,allcolors=cvprblue]{hyperref}
\usepackage{tabularx}
\usepackage{multirow}
\usepackage{booktabs}
\usepackage{xcolor}  
\usepackage[table]{xcolor}
\usepackage{booktabs}
\usepackage{multirow}
\usepackage{array}
\usepackage{subcaption}

\usepackage{tikz}
\usepackage{pgfmath}
\definecolor{monte_carlo}{HTML}{7DE2D1}

\usepackage[accsupp]{axessibility}  

\newcommand{\sys}{SPD}

\newcommand{\numwild}[2]{
    \begin{tikzpicture}[baseline]
        \pgfmathparse{#1 > #2 ? 1 : 0}
        \ifnum\pgfmathresult=0
            \pgfmathsetmacro{\percentdiff}{min(130, 130*(#2-#1)/#2)}
            \pgfmathsetmacro{\intensity}{\percentdiff}
            \fill[monte_carlo!\intensity!white, rounded corners=1]
                (-0.6em, -0.3em) rectangle (2.6em, 1em);
        \fi
        \node[inner sep=0pt] at (1em, 0.7ex) {#1};
    \end{tikzpicture}
}

\newcommand{\squad}{\hspace{0.25em}}
\def\paperID{46310} 
\def\confName{CVPR}
\def\confYear{2026}

\title{Bias Is a Subspace, Not a Coordinate: A Geometric Rethinking of Post‑hoc Debiasing in Vision-Language Models}

\author{
Dachuan Zhao \thanks{\shortstack[l]{Equal contribution. Code available at \href{https://github.com/zhendashen896/SPD}{github.com/zhendashen896/SPD}.}}~$^{1}$\squad
Weiyue Li$^{*1}$\squad
Zhenda Shen$^{*1}$\squad
Yushu Qiu$^{1}$\squad
Bowen Xu$^{1}$\squad
Haoyu Chen$^{1}$\squad
Yongchao Chen\thanks{Corresponding author: yongchaochen@fas.harvard.edu}~$^{2,1}$\\
$^{1}$Harvard University \squad $^{2}$MIT
}

\begin{document}
\maketitle

\input{sections/0_abstract}
\input{sections/1_intro}

\input{sections/2_related}

\input{sections/3_limitation}
\input{sections/4_method}

\input{sections/5_exp}

\input{sections/6_result}
\input{sections/7_discussion}

\section*{Acknowledgement}
We thank the Kempner Institute for the Study of Natural and Artificial Intelligence at Harvard University for computational support.

{
    \small
    \bibliographystyle{ieeenat_fullname}
    \bibliography{main}
}

\input{sections/X_suppl}

\end{document}

%% file: sections/0_abstract.tex
\begin{abstract}
Vision-Language Models (VLMs) have become indispensable for multimodal reasoning, yet their representations often encode and amplify demographic biases, resulting in biased associations and misaligned predictions in downstream tasks. Such behavior undermines fairness and distorts the intended alignment between vision and language. Recent post-hoc approaches attempt to mitigate bias by replacing the most attribute-correlated embedding coordinates with neutral values. However, our systematic analysis reveals three critical limitations of this coordinate-wise approach: feature entanglement, poor cross-dataset generalization, and incomplete bias removal. We find that bias is not localized to a few coordinates but is instead distributed across a few linear subspaces. To address these limitations, we propose \textbf{S}ubspace \textbf{P}rojection \textbf{D}ebiasing (\textbf{\sys{}}), a geometrically principled framework that identifies and removes the entire subspace of linearly decodable bias while reinserting a neutral mean component to preserve semantic fidelity. Extensive experiments across zero-shot classification, text-to-image retrieval, and image generation validate the effectiveness of \sys{}: our method achieves more robust debiasing with an average improvement of 18.5\% across four fairness metrics, while maintaining minimal loss in task performance compared to the best debiasing baseline.
\end{abstract}

%% file: sections/1_intro.tex
\vspace{-0.5cm}
\section{Introduction}
\label{sec:intro}

Vision-Language Models (VLMs) have rapidly become central to modern multimodal AI, powering downstream tasks like image-text retrieval, visual question answering, image captioning, text-to-image generation, cross-domain adaptation, multimodal reasoning, and more~\citep {radford2021learning, li2022blip, hu2024bliva, zhang2026reclaiminglosttextlayers,  lian2023llmgrounded, yang2025med, zhang2024vision}. Their broad generalization and emergent cross-modal alignment make them indispensable foundation models~\citep{alayrac2022flamingo, li2022blip, li2023blip, dai2023instructblip, NEURIPS2023_6dcf277e, xue2024xgen}. However, an expanding body of work shows that VLMs also inherit and amplify demographic and social biases present in large-scale web data~\citep{hall2023visogender, zhao2021understanding, hamidieh2024identifying, janghorbani2023multimodal}. These biases appear as gendered profession associations~\citep{wang2022occupationbias, fraser2024examining, han2025lightfair, hendricks2018women, wu2024evaluating}, racially skewed retrieval rankings~\citep{hall2023visogender, kong2024mitigating, luo2024fairclip}, or stereotypical captions~\citep{zhao2021understanding, cho2023dall, raedler2025the, zhao2017men}, thereby undermining fairness, reliability, and the trustworthiness of model predictions~\citep{zhou-etal-2022-vlstereoset}. Beyond ethical concerns, biased internal representations induce spurious correlations that degrade robustness and cross-domain generalization, as models exploit socially biased features rather than semantically relevant features~\citep{wang2022identifying, geirhos2020shortcut, sagawa2020distributionally, weng2024images, shi2025culturefadesrevealingcultural}. Thus, effective debiasing is essential not only for social responsibility but also for preserving the generalization integrity of multimodal systems.

Existing efforts to debias VLMs follow two lines. Training-based methods fine-tune models to suppress sensitive attributes, which can reduce bias but are computationally heavy, sensitive to hyperparameters, and often tailored to binary attributes with limited transferability. Post-hoc methods operate on frozen embeddings and avoid fine-tuning costs, yet they lack a unified understanding of how bias is represented within VLM embeddings and often rely on inconsistent or fragile assumptions about its underlying geometric structure in high-dimensional spaces. Despite these efforts, how to properly conceptualize and model the geometric structure of bias in multimodal representations remains an open question.

Recent work by~\citet{jung2024sfid} introduces Selective Feature Imputation for Debiasing (SFID), a post-hoc method that identifies the embedding dimensions most predictive of sensitive attributes and replaces them with neutral values derived from low-confidence samples. This coordinate-wise approach is model-agnostic, training-free, and uniformly applicable across VLM components, achieving fairness improvements on multiple benchmarks. Yet SFID's design implicitly assumes bias is localized within a small subset of coordinates, the same dimensions encode a given attribute across different datasets, and replacing these coordinates does not discard semantically relevant information.

However, our systematic reproduction study (\cref{sec:rethink_sfid}) reveals that these assumptions do not hold in practice. We find that the most representative embedding dimensions for different attributes exhibit substantial overlap. This leads to feature entanglement, as removing one attribute’s dimensions unintentionally distorts representations of others. We show that the indices of important dimensions for specific attributes shift across datasets, undermining SFID’s cross-dataset transferability. Finally, we show that replacing the top-$m$ most important coordinates leaves substantial linearly decodable attribute signal, indicating that attribute information is broadly distributed across more dimensions than the imputation process targets~\citep{bengio2013representation}.

Building on these findings, we propose a subspace-projection debiasing framework, \textbf{S}ubspace \textbf{P}rojection \textbf{D}ebiasing (\textbf{\sys{}}), that directly addresses SFID’s limitations by moving beyond its coordinate-wise editing toward a continuous and geometrically principled operation. We explicitly learn bias directions using Iterative Null-space Projection (INLP)~\citep{ravfogel2020null} and project embeddings onto their orthogonal complement, thereby removing attribute-specific components. This approach is more robust and achieves more thorough debiasing than simple coordinate-level interventions. To preserve semantic fidelity, we reinsert a neutral mean from low-confidence samples, which recenters the embeddings without reintroducing attribute-specific variance. This stabilization mitigates overcorrection and improves generalization across datasets and downstream tasks.

We evaluate our framework across three representative downstream tasks: multi-class zero-shot classification, text-to-image retrieval, and text-to-image generation, using multiple VLM backbones. Empirical results show consistently lower demographic-parity gaps and misclassification disparities than SFID and other baselines, while maintaining comparable or higher accuracy and perceptual quality. Our study provides a unified, VLM-training-free, and interpretable approach to post-hoc debiasing of VLMs that supports multi-attribute bias mitigation and advances both fairness and generalization across modalities and tasks.

%% file: sections/2_related.tex
\section{Related works}
\label{sec:related}

\textbf{Training-based Debiasing Methods}\quad  Prior work on VLM debiasing includes some methods that rely on additional training or fine-tuning to suppress sensitive attribute information from learned representations~\citep{berg2022prompt,seth2023dear,hirota2023libra,li2024counterfactualvlm,shen2023finetuning,abdelmagid2024theyre,howard2024socialcounterfactuals,dehdashtian2024fairvlm,ijcai2025p55, li2025achieving}. \citet{berg2022prompt} optimize learnable prompt tokens through an adversarial objective to decorrelate similarity scores from demographic attributes, while~\citet{seth2023dear} introduce a residual correction module that learns additive residuals to remove bias-predictive components from image embeddings. Although these training-based methods can substantially reduce bias, they require heavy hyperparameter tuning and often fail to transfer reliably across datasets. In addition, most are designed for binary attributes and struggle to handle multi-attribute cases in which factors such as race, gender, and age interact. 

\textbf{Training-Free and Post-Hoc Methods}\quad  To avoid the computational cost of retraining, a growing body of work explores training-free or post-hoc methods that operate on frozen models without requiring additional tuning~\citep{chuang2023debiasing,NEURIPS2024_72462f45,ijcai2025p55}. \citet{chuang2023debiasing} identify biased directions in the text embedding space using contrastive prompt pairs and remove them through linear projection, assuming that bias directions are sufficient to separate sensitive demographic attributes from semantics. While effective under designed prompts, this approach is sensitive to prompt selection and struggles to generalize when bias is distributed across more complex or latent subspaces.~\citet{NEURIPS2024_72462f45} extend this idea to the multimodal setting, proposing BEND-VLM, which orthogonalizes image-text embeddings along attribute-specific subspaces at test time to equalize similarity scores across demographic groups. Although broadly applicable across architectures, the method requires per-query optimization during inference, which limits scalability and makes its effectiveness dependent on query semantics and the clarity of underlying attribute dimensions.

\textbf{SFID and Relevant Approaches}\quad  Among post-hoc approaches,~\citet{jung2024sfid} propose SFID, which identifies embedding coordinates most correlated with sensitive attributes and replaces them with neutral statistics estimated from low-confidence samples. This method provides a training-free and computationally efficient solution. However, SFID assumes that bias is localized to specific embedding dimensions and can be mitigated through coordinate-level replacement, which oversimplifies the distributed and entangled structure of multimodal feature spaces.~\citet{zhang2025joint} and~\citet{hirota2024saner} introduce subspace-based neutralization strategies that aim to preserve modality alignment while reducing demographic biases. However, methods such as~\citet{zhang2025joint} rely on attribute supervision, while~\citet{hirota2024saner} are specific to CLIP and thus less transferable to other VLM architectures. Overall, despite the progress of recent post-hoc methods, the field still lacks a unified, training-free framework capable of mitigating bias consistently across modalities while preserving semantic alignment~\citep{lian2026closed}.

%% file: sections/3_limitation.tex
\section{Rethinking SFID}
\label{sec:rethink_sfid}

We conduct a systematic reproduction of SFID~\cite{jung2024sfid} and identify three 
fundamental assumptions underlying its coordinate-wise design. Through 
diagnostic experiments, we show that these assumptions do not consistently hold in practice, motivating our subspace-projection method
\sys{}.

\subsection{SFID Overview and Core Assumptions}
\label{sec:sfid_overview}

Given embeddings $X \in \mathbb{R}^{N \times D}$ from a frozen VLM 
and corresponding attribute labels $y \in \{1, \ldots, C\}$, 
SFID operates in three stages:

\begin{enumerate}
    \item \textbf{Attribute prediction and dimension selection:} A Random Forest classifier~\citep{breiman2001random} is trained to predict the target attribute from $X$. The $m$ dimensions with the highest feature-importance scores are selected as the biased dimension set $\mathcal{S}$.

    \item \textbf{Neutral value estimation:} Samples for which the classifier’s prediction confidence is below a threshold $\tau$ are regarded as low-confidence samples, forming a subset $\mathcal{C}_{\text{low}}$. For each selected dimension $j \in \mathcal{S}$, SFID computes a neutral mean $\bar{x}_{\text{low,j}} = \frac{1}{|\mathcal{C}_{\text{low}}|} \sum_{i \in \mathcal{C}_{\text{low}}} x_{i,j}$ to represent the average feature value among uncertain samples.

    \item \textbf{Feature imputation at inference:} 
During inference, for each query embedding $x_q$, 
the dimensions in $\mathcal{S}$ are replaced with their corresponding 
neutral mean values $\bar{x}_{\text{low,j}}$, while all other dimensions remain unchanged.

\end{enumerate}

This design relies on three assumptions: (A1) different attributes are encoded in disjoint dimensions; (A2) the same dimensions encode a given attribute across datasets; and (A3) bias is concentrated in the top-$m$ dimensions. We test each assumption in the following subsections.

\subsection{Feature Entanglement}
\label{sec:entanglement}

SFID's coordinate-wise replacement strategy assumes that different 
attributes (e.g., gender, race, age) are encoded in disjoint subsets 
of dimensions. Under this assumption, replacing coordinates to debias 
one attribute should not affect the representation of other attributes. 
If this holds, the top-$m$ dimensions selected for predicting different 
attributes should exhibit minimal overlap.

To test this, we train three independent Random Forest classifiers on 
FairFace~\cite{karkkainen2021fairface} CLIP~\cite{radford2021learning} ViT-B/32~\citep{dosovitskiy2020image} 
embeddings ($D{=}512$), one each for Age, Gender, and Race, and extract the 
top-$m{=}100$ dimensions per classifier, following the official SFID repository default setting. We then compute pairwise intersections of these index sets.

\begin{table}[H]
\centering
\footnotesize
\caption{Overlap of top-$m{=}100$ dimensions across age (A), gender (G), and race (R) attributes on FairFace. Higher overlap indicates stronger feature entanglement.}
\label{tab:entanglement}
\begin{tabular}{lcccc}
\toprule
 & \textbf{A $\cap$ G} & \textbf{G $\cap$ R} & \textbf{A $\cap$ R} & \textbf{A $\cap$ G $\cap$ R} \\
\midrule
Overlap & 31 & 37 & 20 & 11 \\
\bottomrule
\end{tabular}
\end{table}

\Cref{tab:entanglement} reveals severe entanglement among attributes, directly contradicting assumption~A1. When SFID replaces 100 dimensions to debias gender, it inevitably disrupts dimensions that are also critical for representing other attributes such as race and age. The results indicate that the same embedding dimensions often carry information about multiple attributes, rather than being cleanly separated. Given that FairFace contains only three labeled attributes, the selected dimensions may additionally encode nuisance factors such as illumination or pose. Consequently, blindly replacing those coordinates as SFID does inject irrelevant noise and simultaneously discards useful semantic information that happens to be co-located in the same dimensions.

\subsection{Cross-Dataset Dimension Drift}
\label{sec:drift}

SFID's transferability relies on the assumption that the dimensions 
identified as most important for a given attribute on 
one dataset should largely align with those on another dataset. If 
true, imputation values $\bar{x}_{\text{low}}$ learned from FairFace 
low-confidence samples can be directly applied to FACET~\cite{gustafson2023facet} embeddings 
by replacing the same coordinate indices. We extract the top-$m$ dimensions most important for gender from FairFace 
and FACET independently, using identical 
Random Forest training procedures, and measure the intersection size. 
As a baseline, randomly selecting two $m$-subsets from $D{=}512$ 
dimensions yields an expected overlap of $m^2/D$ ($\approx 4.9$ for $m{=}50$, $\approx 19.5$ for $m{=}100$).

\begin{table}[H]
\centering
\footnotesize
\caption{Intersection of top-$m$ gender dimensions between FairFace and FACET. Weak alignment shows that direct embedding fails to achieve robust and effective cross-dataset transfer.}
\label{tab:drift}
\begin{tabular}{ccc}
\toprule
$m$ (FairFace) & $m$ (FACET) & Overlap \\
\midrule
50  & 50  & 24 \\
100 & 100 & 40 \\
\bottomrule
\end{tabular}
\end{table}

\setlength{\tabcolsep}{3pt}
\begin{table*}[!ht]
\centering
\caption{Linear-probe accuracy for predicting protected attributes (Race/Gender/Age) from FairFace embeddings after applying each intervention. ``Origin” reports accuracy on the original vanilla embeddings, while the remaining columns report SFID with $m=100$ and \sys{} with $r=1,5,10$ removed subspace directions.}
\label{tab:replace_blocks_k5}
\scriptsize
\begin{tabular}{l c cccc cccc cccc}
\toprule
& & \multicolumn{4}{c}{\textbf{Replace Race}} & \multicolumn{4}{c}{\textbf{Replace Gender}} & \multicolumn{4}{c}{\textbf{Replace Age}} \\
\cmidrule(lr){3-6}\cmidrule(lr){7-10}\cmidrule(lr){11-14}
\textbf{Class} & \textbf{Origin} & \textbf{SFID} & \textbf{SPD($r{=}1$)} & \textbf{SPD($r{=}5$)} & \textbf{SPD($r{=}10$)} & \textbf{SFID} & \textbf{SPD($r{=}1$)} & \textbf{SPD($r{=}5$)} & \textbf{SPD($r{=}10$)} &\textbf{SFID} & \textbf{SPD($r{=}1$)} & \textbf{SPD($r{=}5$)} & \textbf{SPD($r{=}10$)}\\
\midrule
Race   & 0.7144 & 0.7086 & 0.7149 & 0.2745 & 0.1913 & 0.7127 & 0.7140 & 0.7150 & 0.7148 & 0.7118 & 0.7149 & 0.6964 & 0.3080 \\
Gender & 0.9466 & 0.9449 & 0.9466 & 0.9467 & 0.6832 & 0.9404 & 0.9286 & 0.6766 & 0.5925 & 0.9465 & 0.9466 & 0.9397 & 0.5549 \\
Age    & 0.6023 & 0.6023 & 0.6026 & 0.5988 & 0.4745 & 0.5991 & 0.6024 & 0.6009 & 0.5976 & 0.5961 & 0.6023 & 0.3000 & 0.2958 \\
\bottomrule
\end{tabular}
\end{table*}

\Cref{tab:drift} shows weak cross-dataset alignment: when $m{=}50$, the overlap between FairFace and FACET dimensions is about five times higher than random expectation but still well below half of $m$. This indicates that while CLIP stores some shared attribute structure across datasets, many important dimensions still shift when the data distribution changes. Even as $m$ increases to 100, the overlap grows only modestly, confirming genuine index shift rather than a simple re-ranking and thereby violating assumption~A2. Consequently, when SFID transfers low-confidence means from FairFace to other datasets such as FACET, a substantial portion of the replaced coordinates become misaligned, weakening its debiasing effectiveness on unseen data.

\subsection{Incomplete Debiasing}
\label{sec:residual}

SFID assumes that bias is concentrated in the top-$m$ dimensions, such that replacing them should substantially reduce attribute-related information in the modified embeddings. If this assumption holds, replacing the top-$m$ dimensions should at least eliminate the attribute-related signal accessible to linear models, so that a linear classifier achieves accuracy near the random baseline (50\% for binary gender, 14.3\% for 7-class race, 11.1\% for 9-class age). We apply SFID with $m{=}100$ to FairFace embeddings for each attribute 
individually, following the official setup. We then train logistic regression probes on the original and debiased embeddings to assess information removal effectiveness by comparing test accuracies, as shown in \cref{tab:replace_blocks_k5}.
Across all attributes, SFID reduces accuracy by less than 1\% relative to the original embeddings while remaining far above random baselines, indicating that a substantial portion of the target attribute remains linearly decodable from the modified representations. Even replacing 100 out of 512 dimensions (19.5\% of the embedding) barely affects attribute prediction, suggesting that sensitive information is broadly distributed across far more than $m$ coordinates. Prior work on neural representations~\cite{bengio2013representation} has shown that information is often redundantly encoded; SFID’s top-$m$ selection therefore captures only the most concentrated signal. These results suggest that A3 may not strictly hold in practice.

\subsection{Implications for Debiasing Design}
\label{sec:implications}

Our experiments show that all three assumptions underlying SFID's coordinate-wise design do not consistently hold in practice. These failures stem from a fundamental modeling mismatch: SFID treats bias as coordinate-sparse and dataset-invariant, but our findings reveal that bias is subspace-structured, entangled, and distribution-dependent. Consequently, discrete coordinate replacement can be brittle under feature entanglement and distribution shift, and may leave residual linearly decodable bias in our experiments. These observations motivate considering subspace-structured interventions. A principled solution must identify and project onto the null space of \emph{all} attribute-predictive directions, motivating our \sys{} framework in \cref{sec:method}.

%% file: sections/4_method.tex
\section{Methodology}
\label{sec:method}

\subsection{Overview}
\label{sec:overview}
Motivated by the three violations identified in \cref{sec:rethink_sfid}, we propose a subspace-projection framework \sys{} that addresses these limitations. Instead of editing individual coordinates, we model bias as encoded along linear directions in the embedding space and project representations onto the orthogonal complement of a learned attribute-predictive subspace, suppressing linearly decodable attribute information. Because attributes and semantic factors can be partially entangled, this operation can induce collateral utility loss; the projection depth $r$ is therefore treated as a controllable fairness-utility knob.

\begin{figure*}[!ht]
    \centering
    \includegraphics[width=0.9\linewidth,page=1, trim=0 5cm 0 0,
    clip]{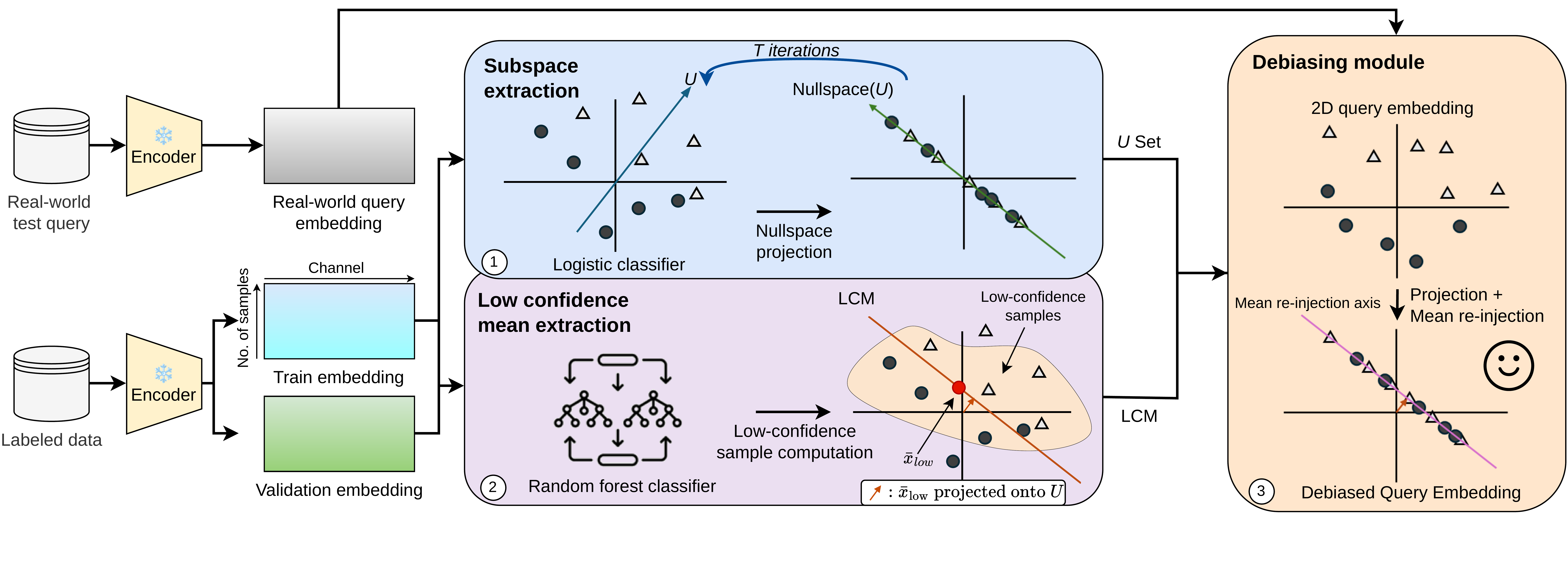}
    \caption{Overview of our framework \sys{} in 2D schematic. (1) We first identify the bias subspace set $U$ via $T$ iterative logistic classifiers, each extracts and removes a bias-predictive direction. (2) We then estimate a neutral mean $\bar{x}_{\text{low}}$ from low-confidence validation samples using a Random Forest. (3) At inference, query embeddings are projected onto the null space of the learned bias subspace and then re-centered by reinjecting the projection of $\bar{x}_{\text{low}}$ along that subspace, yielding debiased representations.}
    \label{fig:pipeline}
\end{figure*}

Given a frozen embedding matrix $X = [x_1, \dots, x_N]^\top \in \mathbb{R}^{N \times D}$ 
and sensitive-attribute labels $y \in \{1, \dots, C\}$, our framework 
consists of three stages: (1)~\emph{Bias subspace identification}, which uses iterative logistic classifiers to extract directions in the embedding space that are predictive of sensitive attributes; (2)~\emph{Subspace projection}, 
which removes embedding components along these directions; and (3)~\emph{Neutral reinjection}, which adds back a low-confidence mean (computed via Random Forest as in SFID) to maintain on-manifold semantics. This procedure is fully post-hoc, requires no 
retraining, and applies to any frozen encoder or decoder. 
\Cref{fig:pipeline} illustrates the complete pipeline.

\subsection{Bias Subspace Identification via INLP}
\label{sec:inlp}
We adopt the INLP framework to extract the linear subspace associated with a sensitive attribute. INLP iteratively identifies and removes the most discriminative directions that enable a linear classifier to predict the sensitive attribute from the embedding. Formally, let \( X^{(0)} = X \) denote the initial embeddings and \( y \) their attribute labels. At iteration \( t \), we train a linear classifier
\begin{equation}
    f^{(t)}(x) = W^{(t)} x + b^{(t)},
\end{equation}
to predict \( y \) from the current embeddings \( X^{(t)} \). For a \(C\)-class attribute, the weight matrix \( W^{(t)} \in \mathbb{R}^{C \times D} \) spans the discriminative subspace that linearly encodes the attribute.
We obtain an orthonormal basis \( U^{(t)} \) for this subspace by performing QR decomposition on \( (W^{(t)})^\top \),
\begin{equation}
    W^{(t)\top} = Q^{(t)} R^{(t)}, \quad U^{(t)} = Q^{(t)\top}_{1:C},
\end{equation}
and define the projection matrix onto its null space as
\begin{equation}
    P^{(t)} = I - \big(U^{(t)}\big)^\top U^{(t)} .
\end{equation}

We then update the embeddings by
\begin{equation}
    X^{(t+1)} = X^{(t)} P^{(t)} .
\end{equation}

This removes the directions most informative of the attribute at iteration \( t \). The process repeats for \( T \) iterations or until the classifier accuracy on \( y \) drops to the random baseline \( 1/C \). The concatenation of all orthogonal bases yields the final bias subspace
\begin{equation}
    U = \big[ U^{(1)}; U^{(2)}; \dots; U^{(T)} \big] \in \mathbb{R}^{d_b \times D},
\end{equation}
where \( d_b = \sum_{t} \mathrm{rank}\!\left(U^{(t)}\right) \). When INLP is run until a linear probe reaches chance (after $T$ iterations), the resulting representation prevents linear attribute recovery beyond chance. In practice, we often remove a limited number of directions $r$ to balance bias suppression and downstream utility. For \(T=1\), INLP reduces to learning a single discriminative direction using a linear classifier. However, real-world embeddings encode sensitive information across multiple correlated directions, and the multi-step INLP procedure provides a more complete and robust removal.

\subsection{Subspace Projection and Neutral Reinjection}
\label{sec:projection}
Once the bias subspace $U \in \mathbb{R}^{d_b \times D}$ has been identified, we remove attribute information by projecting each embedding onto the orthogonal complement of this subspace. For an embedding $x \in \mathbb{R}^D$, the debiased representation is
\begin{equation}\label{eq:proj-only}
    x' = x\,(I - U^\top U).
\end{equation}

Projecting onto the orthogonal complement of $U$ removes the component of $x$ in $span(U)$, thereby reducing linear decodability of the protected attribute along the identified directions. When INLP is run to convergence (until a probe reaches chance), this procedure is designed to make the attribute linearly unpredictable; with a finite number of removed directions $r$, residual linear signal may remain and stronger removal can increase collateral semantic loss. Although orthogonal projection protects linear invariance, it may also remove task-relevant components when attribute and task signals are partially entangled. To stabilize semantics, we add back a \emph{neutral baseline} along the removed subspace. Following SFID~\cite{jung2024sfid}, we employ a Random Forest classifier to estimate prediction confidence for each sensitive attribute. 
For each attribute, we rank samples by confidence and compute the mean embedding of the bottom $\tau$\% (low-confidence) instances, denoted as $\bar{x}_{\text{low}}$, which serves as this neutral baseline. The final representation is
\begin{equation}\label{eq:spd}
    x'' = x' + U^\top\!\big(U\,\bar{x}_{\text{low}}\big).
\end{equation}

The reinjection term is identical for all samples, hence $U x'' = U \bar{x}_{\text{low}}$ is constant across the dataset. Consequently, the operation does not reintroduce attribute-discriminative variability in the removed directions; instead, it recenters the embeddings along the subspace and mitigates off-manifold drift while preserving on-manifold semantics and fairness. This operation generalizes the discrete coordinate replacement of SFID into a continuous, differentiable, and geometrically coherent transformation. Rather than editing individual coordinates, our method removes and then reinstates information along learned bias directions, capturing the distributed structure of bias more effectively.

\vspace{-2pt}
\subsection{Discussion and Properties}
\label{sec:discussion}

Our subspace-projection framework offers several desirable properties: (i) \emph{training-free and efficient} (closed-form operations with lightweight linear probes); (ii) \emph{theoretically grounded} (INLP provably removes crucial linearly decodable bias directions); (iii) \emph{semantically robust} (neutral reinjection mitigates overcorrection); and (iv) \emph{interpretable and controllable} (each learned axis is an explicit, human-inspectable direction). In summary, our method transforms discrete coordinate editing into a principled subspace operation that removes attribute-specific information while preserving semantics, providing a unified, interpretable, and efficient post-hoc debiasing solution for VLMs.

%% file: sections/5_exp.tex
\section{Experiments}
\label{sec:experiments}

We conduct a comprehensive empirical evaluation of SPD to verify its improvements over SFID. The evaluation focuses on two main aspects: (1) how effectively SPD mitigates the limitations identified in \cref{sec:rethink_sfid}, and (2) how it preserves downstream utility across representative VLM tasks. All experiments compare SPD against SFID and other state-of-the-art debiasing baselines.

\subsection{Evaluating SPD on SFID’s Limitations}

\label{sec:debiasing_completeness}

To verify how \sys{} addresses the limitations discussed in \cref{sec:rethink_sfid}, we first evaluate its ability to reduce feature entanglement and achieve more complete debiasing, following the same probe-based setup in \cref{sec:residual}. Specifically, we apply \sys{} to FairFace image embeddings and vary the number of removed subspace directions ($r \in \{1,5,10\}$) for each attribute (race, gender, and age). After projection, we train linear probes on the debiased embeddings to assess whether residual attribute information remains. 

Since \sys{} operates in learned bias directions rather than coordinate indices, it is not directly comparable to SFID in terms of cross-dataset dimension alignment. Instead, we evaluate its robustness to distributional shifts through downstream cross-dataset tasks in \cref{sec:downstream_utility}.

\subsection{Downstream Tasks}
\label{sec:downstream_utility}

Following~\citet{jung2024sfid}, we evaluate performance on three VLM tasks: zero-shot classification, text-to-image retrieval, and text-to-image generation. For each task, we use the same evaluation setup and substitute our debiased encoder for the original model. For SPD, the number of removed subspace directions is set to 5 for all tasks.

\textbf{Baselines}: \textbf{Unmodified model baseline}: The original backbone without any debiasing. \textbf{SFID}~\cite{jung2024sfid}: We use the default configuration where a Random Forest with 100 trees is trained to rank the bias-predictive dimensions by Gini importance, low-confidence samples (threshold $\tau{=}0.7$) are used for imputation, and those dimensions are replaced at inference. \textbf{DeAR}~\citep{seth2023dear}: An adversarial residual debiasing method. We use the default hyperparameters. \textbf{CLIP-clip}~\citep{wang2021gender}: A post-hoc feature pruning method based on mutual information. We first use mutual information to measure how each representation dimension correlates with the sensitive attribute, and then prune out the top-60 most correlated dimensions from the representation. \textbf{Prompt-Debias}~\citep{berg2022prompt} (Prompt-D): A prompt-based debiasing method for text encoders. We use the learned debiasing prompts in the original paper. For retrieval and generation, we report averages over 10 independent runs. For classification, we estimate performance using 1000 bootstrap samples. We report standard deviations in all cases.

\subsubsection{Zero-shot Classification}

{
\setlength{\floatsep}{2pt}
\begin{table*}[!ht]
\footnotesize
\begin{center}
  \caption{Combined experimental results for zero-shot classification (FACET) and text-to-image retrieval (Flickr30K). \textbf{Bold} indicates the best result for each downstream task, while \underline{underline} denotes the second-best result. Improvement (\%) is computed against the baseline of the same model: $(\text{baseline}-\text{method})/\text{baseline}\times100$ for \textit{bias} metrics $\Delta$ DP and Skew@100 (lower is better).}
  \label{tab:zc+retrieval}
  \definecolor{ZS}{HTML}{E8F1FF}   
  \definecolor{RET}{HTML}{FFF2E6}  
  \begin{tabular}{cc*{3}{c}|*{5}{c}}
  \toprule
\multicolumn{2}{c}{\multirow{2}{*}{Model}} 
  & \multicolumn{3}{c}{Zero-shot Multi-class Classification} 
  & \multicolumn{5}{c}{Text-to-Image Retrieval} \\
\multicolumn{2}{c}{} 
  & Accuracy & $\Delta$ DP & $\uparrow$(\%) 
  & R@1 & R@5 & R@10 & Skew@100 & $\uparrow$(\%) \\
\midrule
\multirow{6}{*}{\begin{tabular}[c]{@{}c@{}}CLIP\\ (ResNet50)\end{tabular}}   
& Baseline       & 51.87$\pm$0.58 & 11.08$\pm$0.90 & \textemdash 
                  & 57.24$\pm$0.58 & 81.66$\pm$0.61 & 88.12$\pm$0.56 & 0.1883$\pm$0.0939 & \textemdash \\
& DeAR           & 52.08$\pm$0.63 & 10.04$\pm$0.80 & 9.4 
                  & 57.02$\pm$0.57 & 81.62$\pm$0.76 & 87.95$\pm$0.61 & 0.1817$\pm$0.1207 & 3.5 \\
& CLIP-clip      & 50.73$\pm$0.58 & 10.09$\pm$0.89 & 8.9 
                  & 56.83$\pm$0.43 & 80.99$\pm$0.54 & 87.39$\pm$0.52 & 0.1542$\pm$0.1067 & 18.1 \\
& Prompt-D  & 52.58$\pm$0.56 & 10.37$\pm$0.91 & 6.4 
                  & 57.47$\pm$0.57 & 81.81$\pm$0.75 & 88.23$\pm$0.51 & 0.2030$\pm$0.0971 & -7.8 \\
& SFID           & 50.93$\pm$0.57 & \underline{9.63$\pm$0.86}  & 13.1 
                  & 56.94$\pm$0.51 & 80.89$\pm$0.62 & 87.41$\pm$0.60 & \underline{0.1414$\pm$0.0955} & 24.9 \\
& \textbf{\sys{} (Ours)}     
                  & 51.44$\pm$0.64 & \textbf{9.55$\pm$0.81} & \textbf{13.8}
                  & 56.97$\pm$0.57 & 81.42$\pm$0.66 & 87.85$\pm$0.63 & \textbf{0.1177$\pm$0.0830} & \textbf{37.5} \\
\midrule
\multirow{6}{*}{\begin{tabular}[c]{@{}c@{}}CLIP\\ (ViT-B/32)\end{tabular}}    
& Baseline       & 52.17$\pm$0.58 & 11.60$\pm$0.93 & \textemdash 
                  & 58.91$\pm$0.51 & 83.08$\pm$0.62 & 89.21$\pm$0.48 & 0.1721$\pm$0.0992 & \textemdash \\
& DeAR           & 50.09$\pm$0.45 & 10.37$\pm$0.72 & 10.6 
                  & 59.46$\pm$0.45 & 83.26$\pm$0.66 & 89.23$\pm$0.51 & 0.1387$\pm$0.0912 & 19.4 \\
& CLIP-clip      & 51.56$\pm$0.53 & 10.80$\pm$0.80 & 6.9 
                  & 57.66$\pm$0.73 & 81.80$\pm$0.46 & 87.98$\pm$0.45 & 0.0920$\pm$0.0932 & 46.5 \\
& Prompt-D  & 51.96$\pm$0.53 & 10.56$\pm$0.87 & 9.0 
                  & 58.86$\pm$0.59 & 82.71$\pm$0.62 & 89.08$\pm$0.42 & 0.1496$\pm$0.1097 & 13.1 \\
& SFID           & 52.14$\pm$0.53 & \underline{10.15$\pm$0.85} & 12.5 
                  & 58.53$\pm$0.70 & 82.73$\pm$0.56 & 88.90$\pm$0.56 & \underline{0.0744$\pm$0.0616} & 56.8 \\
& \textbf{\sys{} (Ours)}     
                  & 51.29$\pm$0.52 & \textbf{9.94$\pm$0.76} & \textbf{14.3} 
                  & 59.68$\pm$0.46 & 83.47$\pm$0.54 & 89.35$\pm$0.59 & \textbf{0.0699$\pm$0.0566} & \textbf{59.4} \\
\midrule
\multirow{6}{*}{XVLM}         
& Baseline       & 55.74$\pm$0.48 & 11.72$\pm$0.72 & \textemdash 
                  & 80.77$\pm$0.56 & 96.67$\pm$0.26 & 98.55$\pm$0.23 & 0.2355$\pm$0.1425 & \textemdash \\
& DeAR           & 56.30$\pm$0.52 & 11.26$\pm$0.84 & 3.9 
                  & 78.82$\pm$0.57 & 96.03$\pm$0.39 & 98.17$\pm$0.22 & 0.2066$\pm$0.1667 & 12.3 \\
& CLIP-clip      & 54.52$\pm$0.50 & 9.98$\pm$0.81  & 14.8 
                  & 75.99$\pm$0.54 & 94.77$\pm$0.53 & 97.43$\pm$0.31 & 0.2205$\pm$0.1224 & 6.4 \\
& Prompt-D  & 56.37$\pm$0.48 & 10.35$\pm$0.78 & 11.7 
                  & 79.02$\pm$0.48 & 96.03$\pm$0.36 & 98.24$\pm$0.21 & 0.2355$\pm$0.1658 & 0.0 \\
& SFID           & 53.69$\pm$0.59 & \underline{9.91$\pm$0.92}  & 15.4 
                  & 78.00$\pm$0.46 & 95.67$\pm$0.45 & 98.01$\pm$0.25 & \underline{0.2032$\pm$0.1229} & 13.7 \\
& \textbf{\sys{} (Ours)}     
                  & 54.32$\pm$0.47 & \textbf{9.85$\pm$0.84} & \textbf{16.0} 
                  & 79.13$\pm$0.44 & 96.11$\pm$0.44 & 98.25$\pm$0.20 & \textbf{0.1859$\pm$0.1217} & \textbf{21.1} \\
\bottomrule
\end{tabular}
\end{center}
\end{table*}
}

Zero-shot classification tests whether a VLM can assign an image to the correct class without task-specific fine-tuning: each class is turned into a prompt of the form ``a photo of a/an [CLASS NAME],'' and the predicted label is the one whose text embedding has the highest cosine similarity to the image embedding. Bias arises when accuracy differs across demographic groups within the same class.

We use the FACET benchmark~\citep{gustafson2023facet}, which spans 52 occupation classes with gender, skin tone, and age annotations. We adopt the same class list, prompt template, and preprocessing as SFID. We evaluate this downstream task on each backbone (CLIP~\citep{radford2021learning} with ResNet-50~\citep{he2016deep} and ViT-B/32~\citep{dosovitskiy2020image}, and XVLM~\citep{zeng2021multi}) across all five baselines. Utility is measured by top-1 accuracy. Fairness is measured by 
\begin{equation}\label{eq:dp}
    \Delta DP = \frac{1}{|\mathcal{C}|} \sum_{c \in \mathcal{C}} \big| P(\hat{Y} = c \mid a = 1) - P(\hat{Y} = c \mid a = 0) \big|,
\end{equation}
where $\hat{Y}$ is the predicted class, $c \in \mathcal{C}$ is a task class, and $a \in \{0,1\}$ is the sensitive attribute. This captures the average demographic disparity in per-class prediction rates, i.e., how much more often one group is predicted as class $c$ than another. Lower $\Delta DP$ indicates less disparity.

\subsubsection{Text-to-Image Retrieval}

Given a text query, the model ranks images by cosine similarity between the text and image embeddings, and we evaluate how well the correct image is retrieved. We also check whether the retrieved set is demographically skewed for captions that should be neutral (e.g., returning mostly one gender for ``a person in a suit").

We evaluate text-to-image retrieval on Flickr30K. Following~\citet{wang2021gender}, we modify the ground-truth captions to be gender-neutral, select 1,000 test images. We evaluate this downstream task on each backbone (CLIP ResNet-50, CLIP ViT-B/32, and XVLM) across all five baselines. Utility is measured by Recall@K (K $\in \{1,5,10\}$), the probability that the ground-truth image appears in the top-$K$ retrieved results; higher Recall@K is better. Fairness is measured by Skew@100:
\begin{equation}
    \text{Skew} = \frac{1}{|\mathcal{T}|} \sum_{t \in \mathcal{T}} \max_a \left| \log \left( \hat{p}_a / p_a \right) \right|,
\end{equation}
where $p_a$ is the dataset proportion of attribute $a$ and $\hat{p}_a$ is its proportion among the top $M{=}100$ retrieved images. This term captures, for each sensitive attribute $a$, whether a particular group is retrieved more frequently than expected. Lower Skew indicates more balanced retrieval.

\subsubsection{Text-to-Image Generation}

\begin{table}[!t]
\caption{Experimental results for text-to-image generation.}
\label{tab:generation}
\centering
\renewcommand{\arraystretch}{0.96}

\begin{subtable}[t]{\columnwidth}
\centering
\caption{SDXL}
\footnotesize
\begin{tabular}{lcccc}
\toprule
\multicolumn{1}{c}{} & \multicolumn{3}{c}{Mismatch Rate (Gender prompt)} & Neutral prompt \\
Method & $\vert$M--F$\vert$ & Overall & Composite & $Skew$ \\
\midrule
Baseline               & 3.87$\pm$2.23  & 2.35$\pm$1.22  & 4.42$\pm$2.57  & 83.25 \\
DeAR                   & 89.28$\pm$2.08 & 44.64$\pm$1.04 & 99.81$\pm$2.33 & 99.88 \\
CLIP-clip              & 3.78$\pm$1.88  & 2.11$\pm$1.03  & 4.31$\pm$2.06  & 82.05 \\
Prompt-D          & 39.72$\pm$6.83 & 42.53$\pm$3.85 & 58.49$\pm$3.64 & 82.77 \\
SFID                   & \underline{1.54$\pm$1.14} & \underline{0.84$\pm$0.71} & \underline{1.74$\pm$1.57} & \underline{81.57} \\
\textbf{\sys{} (Ours)} & \textbf{1.48$\pm$0.99} & \textbf{0.78$\pm$0.65} & \textbf{1.67$\pm$1.53} & \textbf{78.66} \\
\bottomrule
\end{tabular}
\end{subtable}

\medskip

\begin{subtable}[t]{\columnwidth}
\centering
\caption{CoDi}
\footnotesize
\begin{tabular}{lcccc}
\toprule
\multicolumn{1}{c}{} & \multicolumn{3}{c}{Mismatch Rate (Gender prompt)} & Neutral prompt \\
Method & $\vert$M--F$\vert$ & Overall & Composite & $Skew$ \\
\midrule
Baseline               & \underline{3.94$\pm$2.71} & 5.54$\pm$2.08  & 6.85$\pm$2.16  & 84.94 \\
DeAR                   & 5.63$\pm$2.84  & 5.42$\pm$1.10  & 8.05$\pm$3.00  & 86.14 \\
CLIP-clip              & 4.73$\pm$2.22  & 5.00$\pm$1.39  & 7.01$\pm$1.53  & 84.58 \\
Prompt-D          & 20.11$\pm$5.15 & 41.99$\pm$2.57 & 46.77$\pm$3.43 & \underline{81.57} \\
SFID                   & 4.70$\pm$1.53 & \underline{2.59$\pm$0.90} & \underline{5.38$\pm$1.44} & 82.77 \\
\textbf{\sys{} (Ours)} & \textbf{3.62$\pm$1.64}  & \textbf{2.53$\pm$0.89} & \textbf{5.26$\pm$1.21} & \textbf{81.20} \\
\bottomrule
\end{tabular}
\end{subtable}
\end{table}

Given a prompt, the model generates an image. We study two failure modes: (i) producing the wrong gender when the prompt specifies one (e.g., generating a woman for ``a photo of a man who works as a/an [profession]''), and (ii) producing a gender-skewed distribution for neutral prompts (e.g., always generating men for ``a photo of a person who works as a/an [profession]''). We use profession-based prompts covering 83 occupations from Bias-in-Bios. For each prompt, we generate 10 images with different seeds.

We evaluate this downstream task on each generative model (Stable Diffusion XL~\citep{podell2023sdxl} and CoDi~\citep{tang2023any}) and compare our SPD method against prior debiasing methods. We first detect the presented gender in each generated image using BLIP-2 with the query ``Does the person look like a male or a female?". For gender-specific p  rompts (``a man" / ``a woman"), we compute the mismatch rate (how often the generated gender disagrees with the requested gender), and report the overall mismatch rate as well as the absolute difference between male-prompt and female-prompt mismatch rates. We also report the composite mismatch rate: 
\begin{equation}
    \text{MRC} = \sqrt{ \text{MR}_O^2 + (\text{MR}_F - \text{MR}_M)^2 }, 
\end{equation}
where $\text{MR}_O$ is the overall mismatch rate and $\text{MR}_F$, $\text{MR}_M$ are the mismatch rates for female- and male-specified prompts. Lower MRC is better.

For neutral prompts (``a person who works as a/an [profession]"), we measure skew toward one gender. Let $N_{p,m}$ and $N_{p,f}$ be the counts of generated male- and female-presenting images for profession $p$, and let $n{=}10$ be the number of generations for that prompt. We define
\begin{equation}
    \text{Skew} = \frac{1}{|\mathcal{P}|} \sum_{p \in \mathcal{P}} \frac{\max(N_{p,m}, N_{p,f})}{n}, 
\end{equation}
which captures whether one gender dominates across professions. Lower Skew indicates more balanced distribution.

%% file: sections/6_result.tex
\section{Results}
\label{sec:results}

\subsection{Effectiveness of \sys{} on SFID's Limitations}
\Cref{tab:replace_blocks_k5} summarizes the results of the evaluation procedure described in \cref{sec:debiasing_completeness}. For the proposed SPD method, when the number of removed subspace directions is $r{=}1$, the accuracy of the target attribute remains almost unchanged, indicating that attribute information is not concentrated in a single direction but distributed across multiple discriminative directions. Increasing $r$ to five substantially reduces the probe accuracy for the target attribute while the non-target attributes change by \(< 1\%\) in most cases. With a moderate projection depth, SPD effectively removes target-attribute information while introducing minimal collateral impact. This indicates that SPD achieves more complete debiasing while substantially alleviating the feature entanglement problem inherent in coordinate-level methods. As $r$ increases to 10, the target accuracies decrease further, but the non-target accuracies also decrease noticeably. This suggests that the lower-ranked directions are more entangled with non-target attributes and semantic signals, where deeper projection removes task-relevant information.

Overall, these results reveal a clear trade-off between debiasing completeness and semantic retention. In practice, choosing a moderate number of projection directions provides a good balance, removing most of the linearly decodable information of the target attribute while largely preserving other attributes and downstream task performance.

\subsection{Better Performance Across Downstream Tasks}

\textbf{Zero-shot Classification task (\cref{tab:zc+retrieval}):} On this task, \sys{} consistently achieves the lowest demographic-parity gaps across all backbones with only minor accuracy movement. Removing the \emph{subspace}, rather than isolated coordinates, reduces disparities more reliably than other baselines, yielding the best fairness-utility trade-off overall. Overall, SPD provides consistent fairness gains and improves the fairness–utility Pareto frontier relative to all baselines.

\textbf{Text-to-image retrieval task (\cref{tab:zc+retrieval}):} \sys{} reduces demographic skew across all backbones, with utility effects that are small and often favorable. In particular, the ViT-B/32 backbone sees simultaneous gains in both fairness and top-1 retrieval, while RN50 attains markedly lower skew with essentially unchanged recall. On XVLM, skew decreases substantially with a moderate recall trade-off.
This pattern indicates that SPD delivers substantial fairness gains for retrieval while keeping retrieval quality.

\textbf{Text-to-image generation task (\cref{tab:generation}):} On this task, \sys{} reduces gender mismatches for gender-specified prompts and produces more balanced outputs for gender-neutral prompts across SDXL and CoDi. Unlike prompt-based or adversarial-residual baselines that could destabilize the generative process, \sys{} maintains generation quality while curbing systematic skew consistently across SDXL and CoDi. These findings confirm that SPD offers the most favorable balance between realism and fairness.

Across diverse datasets and downstream tasks, \sys{} achieves the strongest bias mitigation while preserving semantic consistency. These results suggest that the learned bias subspace exhibits more stable cross-distribution behavior, partially mitigating Assumption~A2 and enabling \sys{} to generalize more robustly across modalities and tasks compared to coordinate-level post-hoc methods.

\subsection{Ablation Study}

{
\setlength{\textfloatsep}{6pt}
\begin{table}[t]
\centering
\footnotesize
\caption{Ablation studies for the effect of neutral re-injection on zero-shot multi-class classification task.}
\label{tab:zs_ablation}
\begin{tabular}{llcc}
\toprule
Model & Method & Accuracy & $\Delta$DP \\
\midrule
\multirow{2}{*}{\begin{tabular}[c]{@{}l@{}}CLIP\\ (ResNet50)\end{tabular}}

& \sys{} proj only & 50.16$\pm$0.67 & \underline{9.61$\pm$0.87} \\
& \sys{} w/ reinject  & 51.44$\pm$0.64 & \textbf{9.55$\pm$0.81} \\
\midrule
\multirow{2}{*}{\begin{tabular}[c]{@{}l@{}}CLIP\\ (ViT-B/32)\end{tabular}}

& \sys{} proj only & 51.04$\pm$0.57 & \textbf{9.91$\pm$0.82} \\
& \sys{} w/ reinject & 51.29$\pm$0.52 & \underline{9.94$\pm$0.76} \\
\midrule
\multirow{2}{*}{XVLM}

& \sys{} proj only & 53.72$\pm$0.48 & \underline{9.87$\pm$0.81}  \\
& \sys{} w/ reinject & 54.32$\pm$0.47 & \textbf{9.85$\pm$0.84}  \\
\bottomrule
\end{tabular}
\end{table}
}
{

\begin{table}[t]
\centering
\footnotesize
\caption{Ablation studies for the effect of $\tau$ for neutral re-injection on zero-shot multi-class classification task.}
\label{tab:zs_ablation_threshold}
\begin{tabular}{llcc}
\toprule
Model & Method & Accuracy & $\Delta$DP \\
\midrule
\multirow{4}{*}{XVLM}

& \sys{}, $\tau = 0.6$ & 53.81 $\pm$0.47 & \textbf{9.84 $\pm$0.84}  \\
& \sys{}, $\tau = 0.7$ & 54.32 $\pm$0.46 & \underline{9.85 $\pm$0.83}  \\
& \sys{}, $\tau = 0.8$ &  54.34 $\pm$0.49 & 9.94 $\pm$0.87 \\
& \sys{}, $\tau = 0.9$ & 54.63 $\pm$0.49 & 10.26 $\pm$0.86 \\
\bottomrule
\end{tabular}
\end{table}
}

We evaluate the neutral reinjection step by comparing our full method \sys{} (``\sys{} w/ reinject," following~\cref{eq:spd}) to a variant without the neutral reinjection (i.e., pure projection or ``\sys{} proj only" by~\cref{eq:proj-only}) on the zero-shot classification task. As shown in~\cref{tab:zs_ablation}, the variant without reinjection yields similar $\Delta$DP but slightly lower accuracy, indicating that the reinjection step helps preserve semantic fidelity. As shown in~\cref{tab:zs_ablation_threshold}, we also experiment with different thresholds $\tau$ for the reinjection and find that $\tau{=}0.7$ achieves stable accuracy while maintaining low $\Delta$DP, and is therefore used throughout. Overall, restoring a neutral, on-manifold component along the removed directions helps preserve semantics without reintroducing significant bias, often accompanied by a slight improvement in performance.

%% file: sections/7_discussion.tex
\section{Conclusion}
\label{sec:conclusion}

We revisit post-hoc debiasing for VLMs and reveal three key systematic flaws in coordinate-level editing approaches, including residual bias leakage, semantic entanglement, and poor cross-distribution generalization. We further find that demographic bias in VLM embeddings is fundamentally subspace-structured rather than coordinate-sparse. Building on this geometric insight, we propose \sys{}, a theoretically grounded and training-free framework that removes controllable attribute-predictive directions and reinjects a neutral mean component. Because protected attributes can correlate with semantic factors, projection depth $r$ governs a fairness-utility tradeoff; across the evaluated datasets and tasks, \sys{} reduces demographic disparity metrics while maintaining competitive task performance.

%% file: sections/X_suppl.tex
\clearpage
\setcounter{page}{1}
\maketitlesupplementary

\section{Qualitative Results for T2I Generation }
\label{sec:appendix2}

\Cref{fig:builder_codi_spd_grid} presents qualitative examples of text-to-image generation results under the neutral prompt ``\textit{a person who works as a film director.}"
The CoDi baseline tends to generate male-presenting images due to training bias, while CoDi + SPD produces more gender-neutral appearances, indicating effective debiasing.

\newcommand{\imgwithlabel}[2]{%
  \begin{minipage}[b]{0.31\columnwidth}
    \centering
    \includegraphics[width=\linewidth]{\detokenize{#1}}\\[-2pt]
    {\small #2}
  \end{minipage}%
}

\begin{figure}[H]
  \centering
  \imgwithlabel{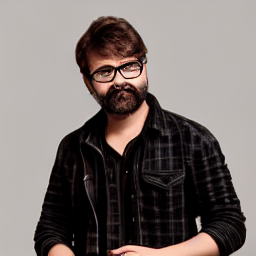}{male}\hspace{6pt}%
  \imgwithlabel{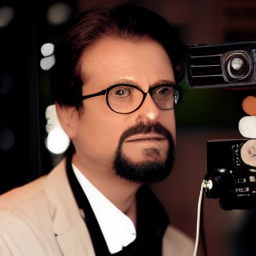}{male}\hspace{6pt}%
  \imgwithlabel{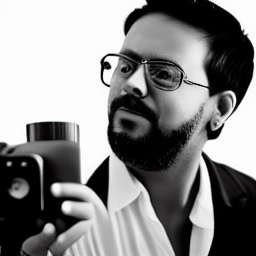}{male}\\[3pt]
  \textbf{CoDi}\\[10pt]

  \imgwithlabel{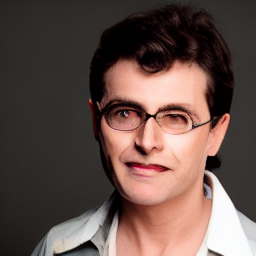}{female}\hspace{6pt}%
  \imgwithlabel{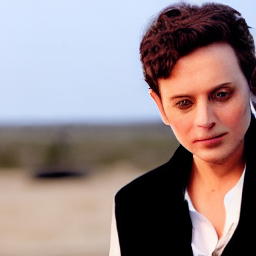}{female}\hspace{6pt}%
  \imgwithlabel{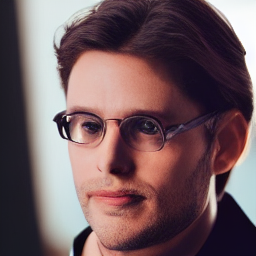}{male}\\[3pt]
  \textbf{CoDi + SPD}

  \caption{Text-to-image generation results for the neutral prompt ``a person who works as a film director."  
  The first row shows CoDi outputs, and the second row shows CoDi with SPD debiasing.  
  Gender labels (``male" / ``female") are automatically assigned using BLIP-2 by asking  
  ``Does the person look like a male or a female?".}
  \label{fig:builder_codi_spd_grid}
\end{figure}

\section{Impact of \texorpdfstring{$r$}{r} on Downstream Tasks}
\label{sec:appendix3}
In SPD, the number of removed bias-predictive directions \(r\) determines a trade-off between debiasing completeness and semantic retention. We first vary \(r\) to study its effect on attribute classification by training linear probes after projection, finding that a moderate projection depth effectively removes target-attribute information while largely preserving semantics. To check whether the optimal setting in downstream tasks aligns with classification, we evaluate text-to-image retrieval on Flickr30K under varying \(r\). As shown in~\cref{tab:retrieval-resnet50},  \(r = 5\) achieves the best balance between retrieval accuracy (R@K) and fairness (Skew@100), consistent with the choice used across downstream evaluations.

\begin{table}[H]
\footnotesize
\begin{center}
  \caption{Text-to-image retrieval (Flickr30K) results for CLIP (ResNet50) with varying $r$.}\label{tab:retrieval-resnet50}
  
  \definecolor{RET}{HTML}{FFF2E6}  
  \begin{tabular}{cc*{4}{c}}
  \toprule
\multicolumn{2}{c}{\multirow{2}{*}{Model}} 
  & \multicolumn{4}{c}{Text-to-Image Retrieval} \\
\multicolumn{2}{c}{} 
  & R@1 & R@5 & R@10 & Skew@100 \\
\midrule

\multirow{6}{*}{\begin{tabular}[c]{@{}c@{}}SPD \end{tabular}}   

& $r=1$  & 57.11$\pm$0.55 & 81.62$\pm$0.74 & 88.05$\pm$0.58 & 0.1613$\pm$0.1025 \\

& $r=3$  & 57.04$\pm$0.73 & 81.65$\pm$0.62 & 88.09$\pm$0.52 & 0.1527$\pm$0.0995 \\

& $r=5$  & 56.97$\pm$0.57 & 81.42$\pm$0.66 & 87.85$\pm$0.63 & 0.1177$\pm$0.0830 \\

& $r=7$  & 55.47$\pm$0.57 & 80.81$\pm$0.75 & 87.13$\pm$0.51 & 0.1165$\pm$0.0971 \\

& $r=10$  & 55.32$\pm$0.65 & 80.62$\pm$0.66 & 86.99$\pm$0.55 & 0.1168$\pm$0.0851 \\

\bottomrule
\end{tabular}
\end{center}
\end{table}

\section{Computing Resource}
\label{sec:appendix_compute}

\Cref{tab:compute_resources} presents the hardware configuration and runtime of each module used in our experiments.
The reported times include Random Forest training for the subsequent projection-direction extraction.
Overall, SPD adds limited computational cost during inference while maintaining efficiency across all tested backbones.

\begin{table}[h]
\small
\centering
\caption{Compute resources used for experiments.}
\label{tab:compute_resources}
\begin{tabularx}{\linewidth}{@{}l X@{}}
\toprule
\textbf{Component} & \textbf{Details} \\
\midrule
CPU & Intel Xeon Gold 6430 \\
Numbers of CPU cores used& 32\\
GPU & NVIDIA RTX A100\\
Numbers of GPU used& 4\\
\midrule
(CLIP ViTB-32 Image Encoder) & \\
\quad Training RandomForest & 310.47s \\
\quad  Direction Extraction & 37.36s \\
(CLIP RN50 Image Encoder) & \\
\quad  RandomForest Training& 466.52s \\
\quad  Projection Direction Extraction & 35.37s \\
\midrule
(CLIP ViTB-32 Text Encoder) & \\
\quad Training RandomForest & 1571.34s \\
\quad  Prodection Direction Extraction & 93.77s \\
(CLIP RN50 Text Encoder) & \\
\quad  RandomForest Training& 2567.06s \\
\quad  Projection Direction Extraction & 146.31s \\
\midrule
(CoDi Text Encoder) &\\
\quad Training RandomForest  & 62.71s \\
\quad Projection Direction Extraction & 3.73s \\
(CoDi Image Dncoder) &\\
\quad Training RandomForest  & 197.33s \\
\quad Projection Direction Extraction & 52.63s \\
Inference on CoDi with SPD & 403.62s / 249 prompts \\
\bottomrule
\end{tabularx}
\end{table}